\documentclass[letterpaper, 10 pt, conference]{ieeeconf}  
\IEEEoverridecommandlockouts                              
\usepackage{graphicx}
\usepackage{times}
\usepackage{amsmath}
\usepackage{amssymb}
\usepackage[utf8]{inputenc}
\usepackage[colorlinks=false,plainpages=false,pdfborder={0 0 0},backref=false,bookmarks=false]{hyperref}
\usepackage{rgap_notation}
\usepackage{rgap_header}

\newcommand*{\field}[1]{\mathbb{\MakeUppercase{#1}}}		
\newcommand*{\set}[1]{{\mathcal{\MakeUppercase{#1}}}}			

\newcommand*{\sequence}[1]{\mathrm{\MakeUppercase{#1}}}

\newcommand*{\R}{\field{R}} 

\renewcommand{\vec}[1]{{\boldsymbol{\mathbf{#1}}}}
\newcommand*{\mat}[1]{\vec{\MakeUppercase{#1}}}
\newcommand*{\eye}{\mat{I}}							
\newcommand*{\transpose}{\mathsf{T}}

\newcommand*{\observation}{y} 					
\newcommand*{\observations}{\vec{\observation}}
\newcommand*{\obsNoise}{\nu}

\newcommand*{\gpMeanFunction}{m}

\newcommand*{\parameters}{\vec{x}} 
\newcommand*{\ParamSpace}{\set{X}} 

\newcommand*{\state}{s}
\newcommand*{\StateSpace}{\set{S}}
\newcommand*{\action}{a}

\newcommand*{\cost}{c}
\newcommand*{\Cost}{C}
\newcommand*{\trajectory}{\sequence{\state}}
\newcommand*{\stimulus}{v}
\newcommand*{\stimuli}{\sequence{\stimulus}}
\newcommand*{\iReward}{r}       

\newcommand*{\normal}{\mathcal{N}}					

\newcommand*{\af}{h} 						
\newcommand*{\Sspace}{\set{S}} 				

\newcommand*{\nObs}{n}

\newcommand*{\nTrajectories}{M}
\newcommand*{\nFeatures}{m}

\newcommand*{\anyfunction}{g}

\newcommand*{\iid}{i.i.d.\xspace}

\title{\LARGE \bf
Bayesian Optimisation for Robust Model Predictive Control \\
under Model Parameter Uncertainty 
} 

\author{Rel Guzman, Rafael Oliveira, and Fabio Ramos
\thanks{The authors are with the School of Computer Science, the University of Sydney, Australia. Rafael Oliveira is also with the Australian Research Council's centre in Data Analytics for Resources and Environments (DARE) and the University of Sydney's Brain and Mind Centre, and Fabio Ramos is also with NVIDIA, USA. { \tt \{rel.guzmanapaza, rafael.oliveira, fabio.ramos\}@sydney.edu.au}\textbf{•}
}}

\begin{document}

\maketitle

\begin{abstract}
We propose an adaptive optimisation approach for tuning stochastic model predictive control (MPC) hyper-parameters while jointly estimating probability distributions of the transition model parameters based on performance rewards. In particular, we develop a Bayesian optimisation (BO) algorithm with a heteroscedastic noise model to deal with varying noise across the MPC hyper-parameter and dynamics model parameter spaces. Typical homoscedastic noise models are unrealistic for tuning MPC since stochastic controllers are inherently noisy, and the level of noise is affected by their hyper-parameter settings. We evaluate the proposed optimisation algorithm in simulated control and robotics tasks where we jointly infer control and dynamics parameters. Experimental results demonstrate that our approach leads to higher cumulative rewards and more stable controllers.
\end{abstract}

\section{Introduction}

Stochastic model predictive control (MPC) methods have been successfully used in many applications, from steady-state control to path planning in robotics~\cite{Pravitra2020}. These methods rely on a model of the system dynamics to obtain optimal control strategies, or actions, by weight-averaging sampled trajectories at each time step. There have been several variations of these methods, including path integral and cross-entropy approaches \cite{Williams2018a}. Unfortunately, because stochastic MPC relies on a dynamics model, its robustness degrades when such a model is different from the true dynamics. The controller could misrepresent the environment, wrongly optimising a performance function on an alternative dynamical system, which might not translate to optimal actions on the real system. The broad term for this issue is {\em reality gap}, which appears in sim-to-real transfer in the context of robotic simulators \cite{Peng2017,Ramos2019} and system identification as typical cases. Despite the reliance on an accurate model, model-based control and model-based reinforcement learning typically require fewer interactions with the environment, which makes them more suitable for real-world robotics applications than model-free approaches~\cite{Wang2019}.

In parallel developments, Bayesian optimisation (BO) as a black-box optimisation method \cite{Shahriari2016} has been applied to control and robotics to optimise hyper-parameters in environment control \cite{Antonova2019, Fiducioso2019} and real robotic manipulators \cite{Dries2017}. The conventional homoscedastic BO approach, however, is based on the strict assumption of constant observation noise. Heteroscedastic noise models can better quantify uncertainty in practical control problems by considering input-dependent noise. In \cite{Guzman2020}, varying noise levels were discovered for a stochastic MPC hyper-parameter $\sigma_\epsilon$ that determines control variance in a model predictive path integral (MPPI) \cite{Williams2018a} controller for benchmark control problems.

As a way to deal with the reality gap, domain randomisation has been applied to randomise simulators to expose the robot to different scenarios instead of learning controllers on a single simulated scenario \cite{Tobin2017}. Domain randomisation could be applied at any component of the reality gap, and it has been used to simulate environments \cite{Muratore2021} and to optimise the robot model parameters \cite{Ramos2019,Lee2020}. 

In this paper, we address the reality gap problem in stochastic MPC by deriving a BO framework to learn controllers which are robust under model uncertainty. Our framework deals with robustness since it does not have to be precise regarding the dynamics model parameters. We perform adaptive domain randomisation to automatically estimate probability distributions over model parameters solely based on task performance data while jointly optimising the controller's hyper-parameters. To account for non-homogeneous response noise, we extend the framework in \cite{Guzman2020} which employs heteroscedastic BO to tune MPPI controllers. Our contributions can be summarised as follows:
\begin{itemize}
\setlength{\itemsep}{1pt}
  \setlength{\parskip}{0pt}
  \setlength{\parsep}{0pt}
\item a framework for tuning stochastic MPC while jointly estimating probability distributions of the transition model parameters, which is based only on observing rewards;
\item an analysis on whether capturing the uncertainty of dynamic parameters leads to better performance;
\item experimental results on benchmark simulated classic control and robotic problems.
\end{itemize}

\section{Background}
\label{sec:background}

\subsection{Stochastic Model Predictive Control}
\label{subsec:mppi_definition}

Model predictive control (MPC) consists of optimising robot actions over a horizon $T$ based on the idea of solving inner cost optimisation problems over predicted trajectories. MPC returns a next optimal action $a^*$ that is sent to the system actuators. A stochastic MPC method models disturbances as random variables. At each time step $t$, stochastic MPC generates sequences of perturbed actions $V_t = \{\action_i^* + \epsilon_i\}_{i=t}^{t+T}$ where $\epsilon_i \sim \normal(0, \sigma_\epsilon^2)$, based on a roll-over sequence of optimal actions $\{\action_i^*\}_{i=t}^{t+T}$ that start as 0. Each action results in a state produced by a transition or dynamics model $s_{t+1} = f\left(s_{t}, a_{t}\right)$. Action sequences result in a state trajectory $\trajectory_t = \{\state_{t+i}\}_{i=1}^T$. Each one has a cost determined by a cumulative function $\Cost$ with instant cost $c$ and terminal cost $q$:
\begin{align}
\Cost(\trajectory_t)=q\left(s_{t+T}\right) + \sum_{i=1}^{T-1} \cost(s_{t+i})~.
\end{align}
A stochastic method known as model predictive path integral (MPPI) and its variations \cite{Williams2018a} provide optimal actions for the entire horizon working under an information-theoretic approach. After $M$ rollouts, MPPI updates a sequence of optimal actions and weights:
\begin{equation}
\action_{i}^* \leftarrow \action_i^* + \sum_{j=1}^{\nTrajectories} w(\stimuli_t^{j})\epsilon_i^{j}
\,,
\label{eq:mppi_action}
\end{equation}
\vspace{-3mm}
\begin{equation}
w(\stimuli_t) = \frac{1}{\eta} \exp\left(-\frac{1}{\lambda}\left(\Cost(\trajectory_t) + \frac{\lambda}{\sigma_\epsilon^2} \sum_{i=t}^{t+T}\action_{i}^*\cdot\stimulus_{i}\right)\right)\,,
\label{eq:mppi_eq}
\end{equation}
where $j \in \{1,\ldots, \nTrajectories\}$ and $\eta$ is a normalisation constant. For the hyper-parameter temperature $\lambda \in \R^+$, $\lambda \to 0$ leads to a single trajectory having higher probability of occurrence, and there is also the control variance $\sigma_\epsilon^2$ that results in more varying and forceful actions when it increases \cite{Williams2018a}. In this way, both $\lambda$ and $\sigma_\epsilon$ control exploration and exploitation of trajectories. The other hyper-parameters are horizon $T$ and number of roll-outs $M$. 

\subsection{Gaussian Processes}
\label{sec:gp}
Gaussian processes are non-parametric probabilistic models to approximate functions \cite{Rasmussen2006}. Considering a function $g:\ParamSpace\to\R$, GPs are specified by a mean function $\vec{\gpMeanFunction} = m(\mat\parameters)$ and a covariance function $k:\ParamSpace\times\ParamSpace\to\R$. Given a design matrix of $n$ points $\vec{X}$, function evaluations $g(\vec{X}) = [\anyfunction(\parameters_1),\dots,\anyfunction(\parameters_\nObs)]^\transpose$ where $\{\parameters_i\}_{i=1}^\nObs \subset\ParamSpace$ are Gaussian:
\begin{equation}
\anyfunction(\mat\parameters) \sim \normal(\vec{\gpMeanFunction},\mat K)~,
\label{eq:gp1}
\end{equation}
where $\mat K$ is a $\nObs$-by-$\nObs$ covariance matrix $[\mat K]_{i,j} = k(\parameters_i,\parameters_j)$. By denoting a test point as $\vec{x}_*$, we condition $\anyfunction_* = \anyfunction(\parameters_*)$ on the observations $\vec{y} = g(\vec{X}) + \vec\obsNoise$, where $\vec\obsNoise \sim \normal(\vec 0, \mat\Sigma_\obsNoise)$ represents observation noise, producing a Gaussian posterior distribution $\anyfunction_*|\vec{X}, \observations, \vec{x}_* \sim \normal(\bar{g}_*,\operatorname{var}(g_*))$. The posterior infers $g$ at unobserved locations and has the closed form:
\begin{align}
\bar{g}_* &= \gpMeanFunction(\parameters_*) + k(\parameters_*, \mat\parameters)(\mat K + \mat\Sigma_\obsNoise)^{-1}(\observations-m(\mat\parameters))\\
\operatorname{var}(g_*) &= k(\parameters_*, \parameters_*) - k(\parameters_*, \mat\parameters)(\mat K + \mat\Sigma_\obsNoise)^{-1}k(\mat\parameters, \parameters_*)~,
\label{eq:posteriors}
\end{align}
Under a homoscedastic noise assumption, $\mat\Sigma_\obsNoise = \sigma^2_\obsNoise\eye$. In this paper, however, we assume a heteroscedastic, i.e.\ input-dependent, noise formulation where $[\mat\Sigma_\obsNoise]_{ij} = k_\obsNoise(\parameters_i,\parameters_j)$ and $k_\obsNoise:\ParamSpace\times\ParamSpace\to\R$ is a positive-definite covariance function.

\subsection{Bayesian Optimisation}
We use BO \cite{Shahriari2016} as it has been applied in robotics and control to optimise expensive black-box functions. Given a search space $\ParamSpace$, BO commonly uses a Gaussian process (GP) (\autoref{sec:gp}) 
as a surrogate model $\mathcal{M}$ to internally approximate an objective function $\anyfunction:\ParamSpace \to \mathbb{R}$, which is commonly a reward maximisation problem $\parameters^* \in \argmax_{\parameters \in \ParamSpace} \anyfunction(\parameters)$.
Given a set of collected observations $\mathcal{D}_{1:t}$, BO constructs the surrogate model that provides a posterior distribution over the objective function $g$. This posterior is used to construct the acquisition function $h$, which measures both performance and uncertainty of unexplored points. The next step is optimising the acquisition function, obtaining a sample $(\vec{x}_t, y_t)$. We use the acquisition function known as upper confidence bound (UCB) \cite{Cox1992} due to its simplicity in balancing exploration and exploitation with the balance factor $\delta > 0$. A high value of $\delta$ leads to more exploration: 
\begin{equation}
\af(\vec{x}):=\mu(\vec{x}) + \delta \sigma(\vec{x})~, \quad
\vec{x}_t\in\argmax_{\parameters \in \Sspace} \af(\vec{x}, \mathcal{M}, \mathcal{D}_{1:t})~,
\label{eq:ucb_problem}
\end{equation}

where $\mu(\vec{x})$ is the posterior mean and $\sigma(\vec{x})$ is the posterior variance of the objective function $g(\vec{x})$. Both are obtained with a surrogate model $\mathcal{M}$ trained with $\mathcal{D}_{1:t}$. The next sampling point $\vec{x}_t$ is obtained by solving \autoref{eq:ucb_problem}.

\section{Methodology}
\label{sec:method}
This section describes our heteroscedastic Bayesian optimisation method for adaptive MPC. We start with a description of the transition model and the adaptation problem we consider and then proceed with adaptive MPC and our approach to model heteroscedastic noise.

\subsection{Transition Model and Adaptation}

We use a dynamics model with model parameters $\theta$ and a deterministic state transition model $s_{t+1} = f\left(s_{t}, a_{t}, \theta \right)$ where a given state $s_{t+1}$ depends only on the previous state $s_t$ and the applied action $a_t$. The system state $\state \in \StateSpace$ lies in a continuous state space $\StateSpace$, and $a \in \mathcal{A}$ is an action specified by a discrete action space. Upon the execution of an action, the system transitions to the next state $s_{t+1}$, producing an instantaneous reward $\iReward_t : \mathcal{S} \times \mathcal{A} \to \R$ that measures the system performance at a given state and action.

\begin{figure}[h]
    \centering
    \subfloat[Gamma distributions]{\includegraphics[width=0.495\columnwidth]{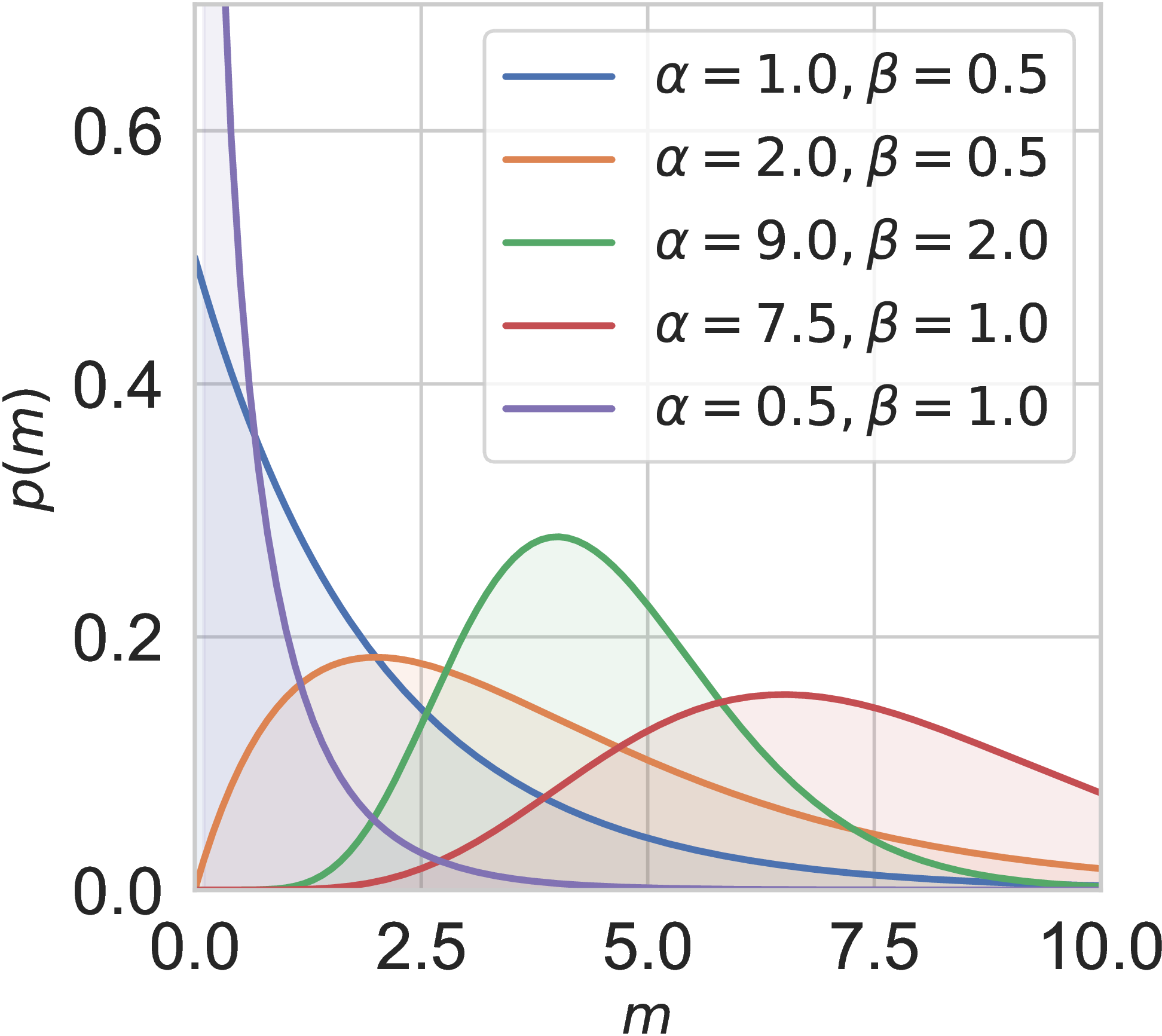}\label{fig:disgamma}}
    \subfloat[Beta distributions]{\includegraphics[width=0.495\columnwidth]{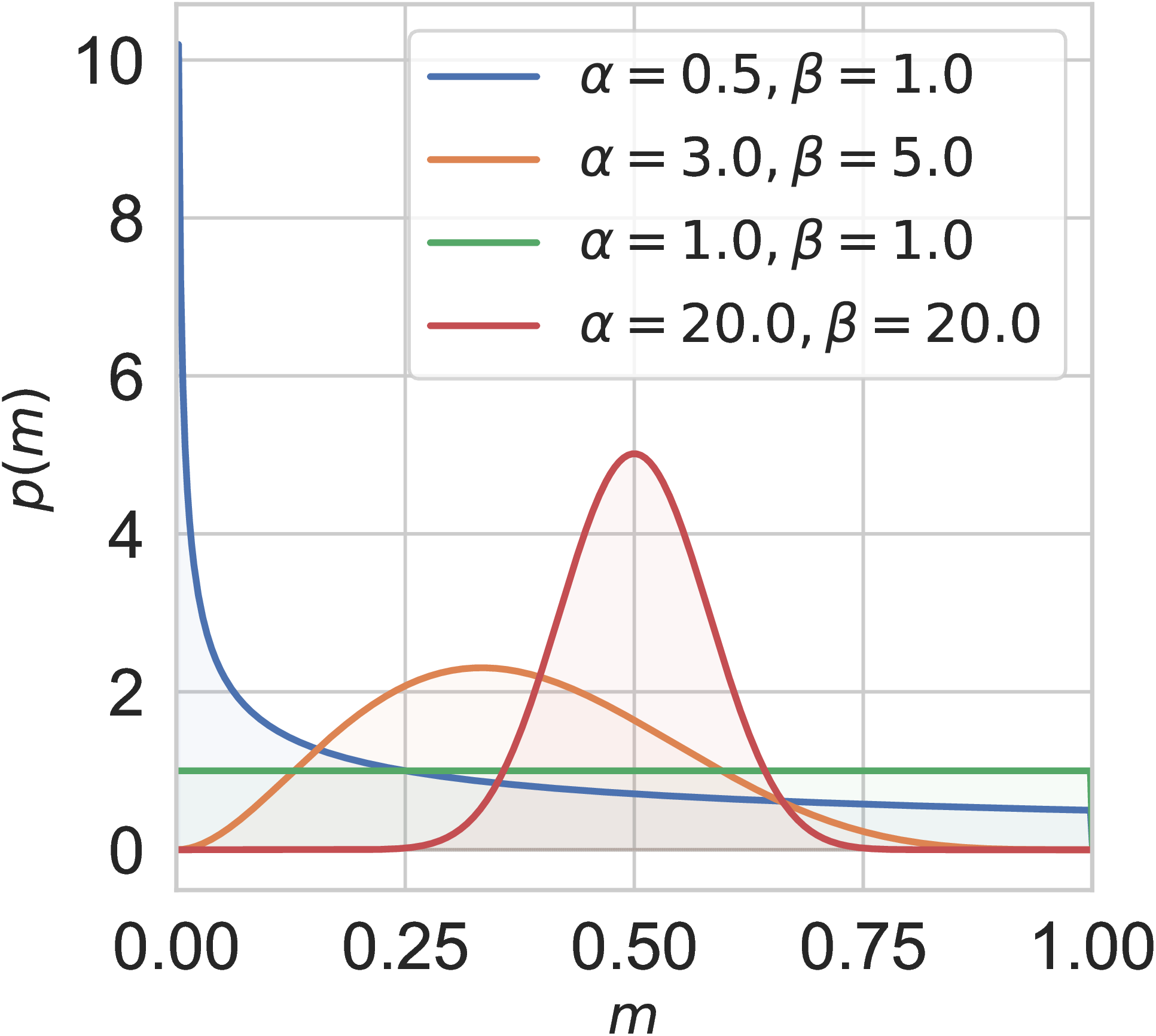}\label{fig:disbeta}}
    \caption{Examples of distributions for different parameters.}
    \label{fig:distribs}
\end{figure}

While stochastic MPC optimises actions, we add model parameter randomisation at each trajectory rollout in order to adapt the controller to the transition model in real-time. For instance, in the case of the Pendulum problem, the parameter mass $\theta=m$ could be beta-distributed, $m \sim \operatorname{Beta}(\alpha, \beta)$, and we would obtain mass distributions as in \autoref{fig:disbeta}. In such a case, we would have an adaptive transition model $s_{t+1} = f\left(s_{t}, a_{t}, m\right)$ and the objective would be to adapt a prior mass distribution $p(m)$ to a robust mass probability distribution with expected value around the true mass. 

\begin{figure}[h]
    \centering
    \includegraphics[width=0.43\columnwidth]{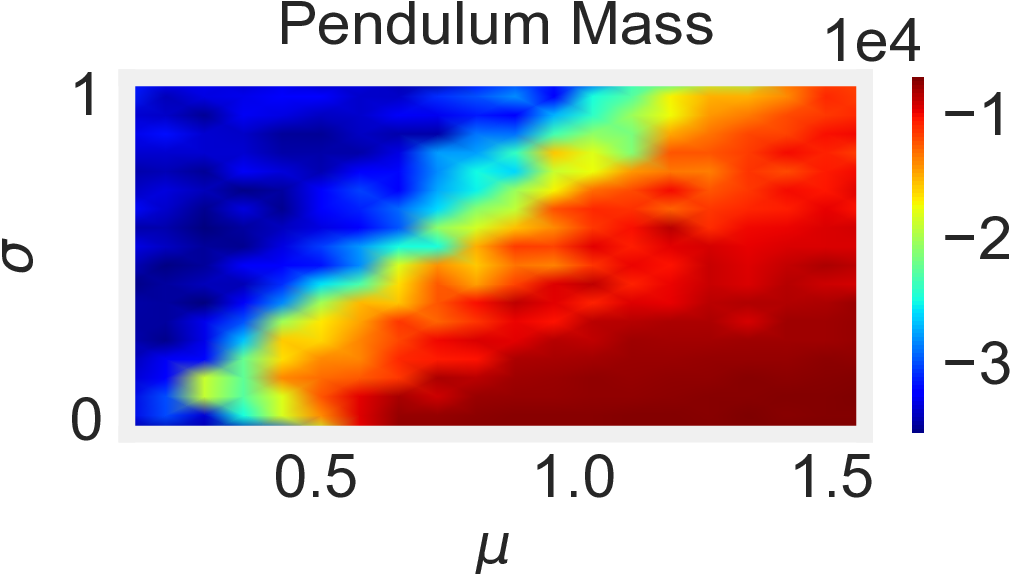}\hspace{1.9mm}
    \includegraphics[width=0.43\columnwidth]{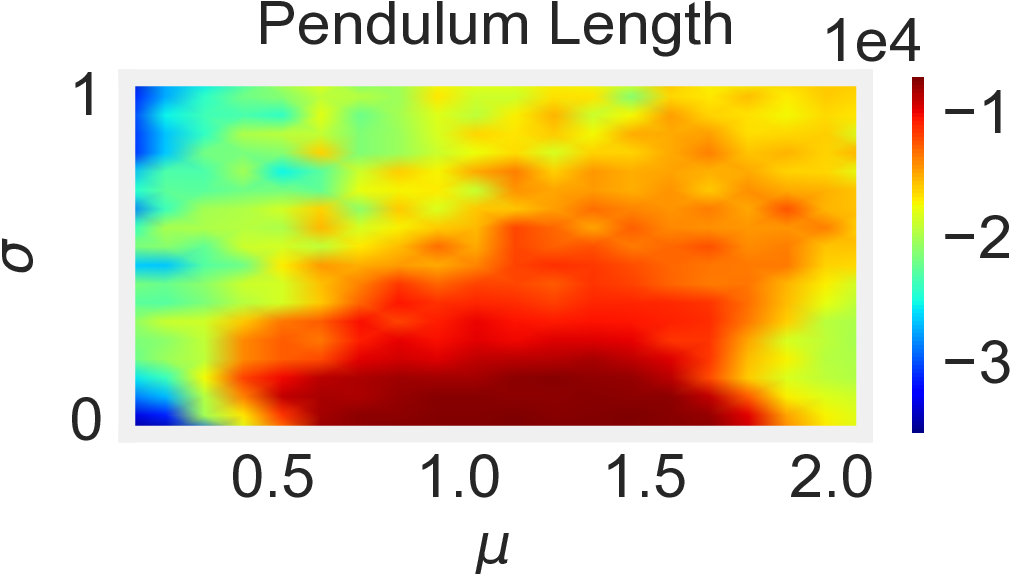}
    \caption{Rewards for model parameter distributions.}
    \label{fig:pendulum_ditribs}
\end{figure}

To handle model uncertainty, we consider model parameters as random variables with values within a given range. The closer the range is from the true parameter, the higher the reward as it appears in \autoref{fig:pendulum_ditribs}, which shows cumulative rewards for possible distributions of the pendulum mass and length parameterised by $\mu$ and $\sigma$. The red regions present the highest cumulative rewards. We estimate these optimal parameter distributions via Bayesian posterior inference.

We define a random vector of model parameters $\boldsymbol{\theta}$ and adapt them to the heteroscedastic MPC controller. Each model parameter follows a probability distribution parameterised by $\boldsymbol{\psi}$: 
\begin{equation}
\boldsymbol{\theta} \sim p_{\boldsymbol{\theta}}(\boldsymbol{\theta}; \boldsymbol{\psi}), \qquad \mathbf{s}_{t+1} = f\left(\mathbf{s}_{t}, \mathbf{a}_t + \boldsymbol{\epsilon}_t, \boldsymbol{\theta} \right)~,
\end{equation}
where $\mathbf{s}_{t}$ is a vector of states obtained at time $t$, and $\mathbf{a}_t + \boldsymbol{\epsilon}_t$ is a vector of perturbed actions. $\boldsymbol{\theta}$ become transition model inputs. Finally, optimal actions found by MPC are sent to the system using the adaptive transition model $f$.

\subsection{Adaptive Model Predictive Control}

\begin{figure}[ht]
    \centering
    \includegraphics[width=0.485\columnwidth]{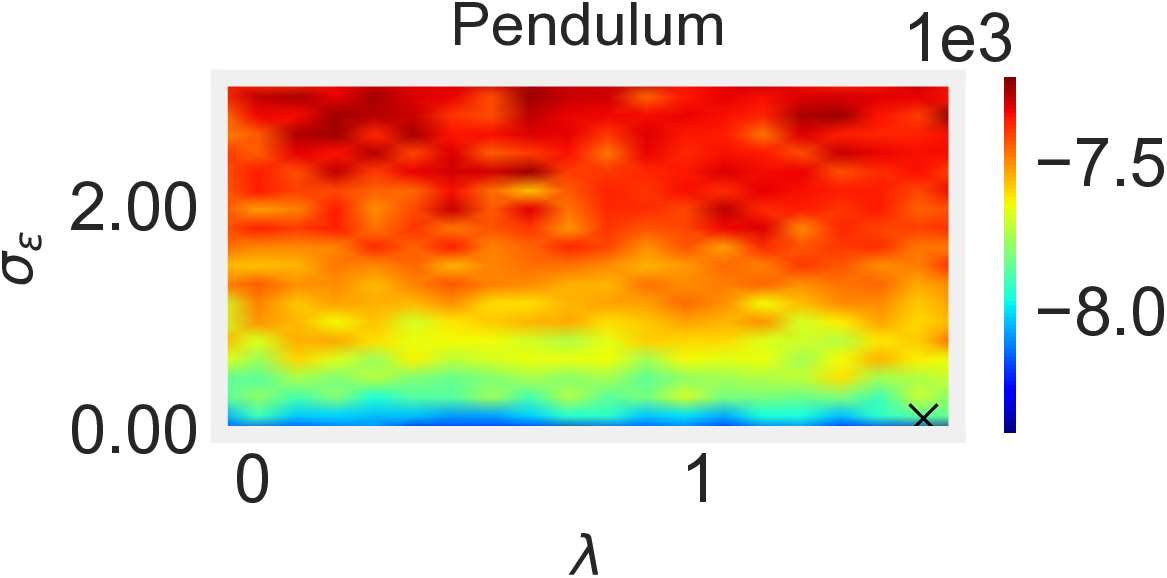} \hspace{0.3mm}
    \includegraphics[width=0.485\columnwidth]{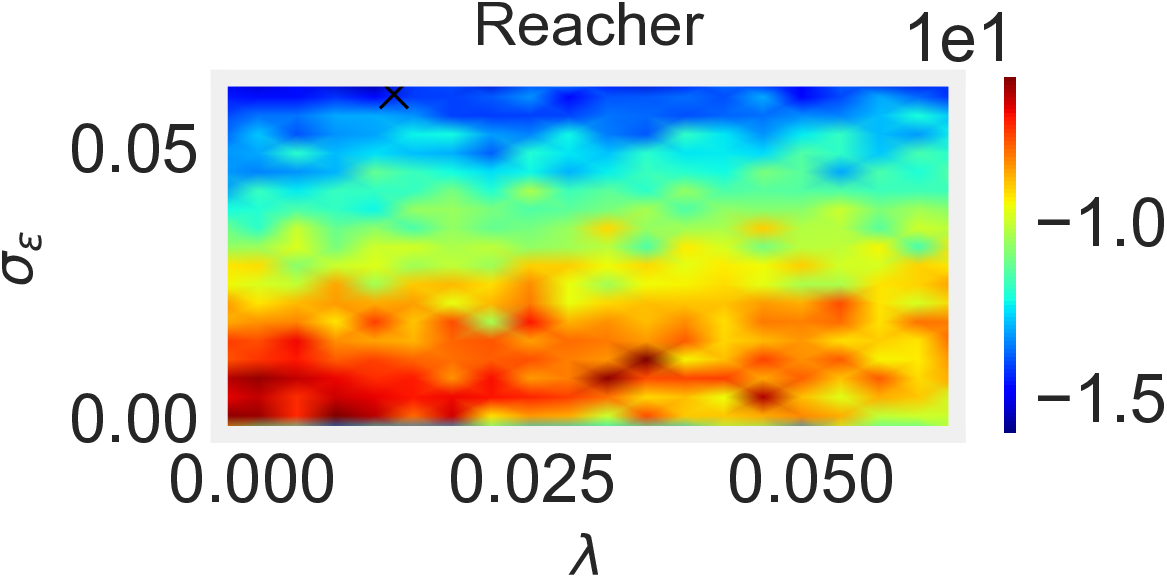}
    \caption{Rewards for MPPI hyper-parameter combinations.}
    \label{fig:2dmppi}
\end{figure}

We do not directly aim at finding parameters that match the observed dynamics as we would do in a system identification approach \cite{Romeres2016}. Instead, we adapt model parameter distributions to the controller, which means those resulting distributions may not be close to their true values. We make sure that we are using the right MPC hyper-parameters by adapting them to the transition model parameters. In the case of MPPI \cite{Williams2018a}, the hyper-parameters are the temperature $\lambda$, noise $\sigma_\epsilon$, horizon $T$, and the number of rollouts $M$ as described in \autoref{subsec:mppi_definition}. They can be collectively described as $\boldsymbol{\phi}$. In this work, we only work with $\lambda$ and $\sigma_\epsilon$. \autoref{fig:2dmppi} shows cumulative rewards for grid search hyper-parameter combinations of Pendulum and Reacher simulators where optimal regions are in red.

We optimise the episodic cumulative reward $R$ by jointly optimising $\boldsymbol{\psi}$, which parameterises the model parameters $\theta$, and the controller hyper-parameters $\boldsymbol{\phi}$. The optimisation problem is described as follows:
\begin{equation}
\boldsymbol{\psi}^{\star}, \boldsymbol{\phi}^{\star}=\argmax_{\{\boldsymbol{\psi}, \boldsymbol{\phi}\}}\, R(\boldsymbol{\psi}, \boldsymbol{\phi})~.
\label{eq:optproblem}
\end{equation}
We define $R = \sum_{i=1}^{n_s} r_i$, where $n_s$ is the number of time steps of an episode. We aim to maximise the expected cumulative reward $g := \mathbb{E}[R]$ of the controller under the real transition dynamics, which is unknown. We approximate the expected cumulative reward $g$ empirically by averaging over $n_e$ episodes.

In order to handle noisy heteroscedastic observations, we define $\parameters = \{\boldsymbol{\psi}, \boldsymbol{\phi}\}$ as a variable to optimise the cumulative reward as performance measure $\anyfunction$ with BO posterior inference $\anyfunction_*|\vec{X}, \observations, \vec{x}_* \sim \normal(\bar{g}_*,\operatorname{cov}(g_*))$ where $\vec{X}$ and $\observations$ is a set of hyper-parameter observations and $(\vec{x}_*, \anyfunction_*)$ are the hyper-parameter setting and reward sampled by BO. This way, we are able to find optimal MPC hyper-parameters adapted to the transition model.

\subsection{Parametric Heteroscedastic Noise Model}
\label{sec:heteroparam}

Cumulative rewards as a function of hyper-parameters for stochastic MPC present varying noise \cite{Guzman2020}. For that reason, a heteroscedastic noise model for BO optimisation should be considered. We use a heteroscedastic noise model $\mathbf{\Sigma}_\obsNoise$ for the GP defined in \autoref{eq:gp1}. To define $\mathbf{\Sigma}_\obsNoise$, we assume that episodes are executed independently, which means that $k_\obsNoise(\parameters,\parameters') = 0$ for $\parameters \neq \parameters'$. Then, we model only $k_\obsNoise(\parameters, \parameters) = \sigma_\obsNoise^2(\parameters)$. Previously, \cite{Guzman2020} defined the parametric noise model:
\begin{align}
    \sigma_\obsNoise(\parameters) &= z \cdot \exp\left(\vec{\beta}^\transpose\vec{\rho}(\parameters)\right) + \zeta~,
    \label{eq:noise-model}
\end{align}
where $\vec\beta\in \R^\nFeatures$, $\zeta \geq 0$ and $\vec{\rho}: \ParamSpace\to\R^\nFeatures$ is a feature map. Small values of $z$ would produce functions that are closer to $\hat{g}$, and when $z$ is large, the model will account for more outliers. The offset term $\zeta$ represents minimum homoscedastic noise.

The exponential term ensures that $k_\obsNoise$ is always positive definite and includes a generalised linear model $\vec{\beta}^\transpose\vec{\rho}(\parameters)$ that fits the expected cumulative reward. The feature map $\vec{\rho}$ could be polynomial or kernel-based. The degree of the polynomial depends on the cumulative reward function to fit. In our experiments, we used a polynomial of degree 10, which was sufficient to model the noise we observed in practice. 

\begin{figure}[t]
\centering
\includegraphics[width=0.420\columnwidth]{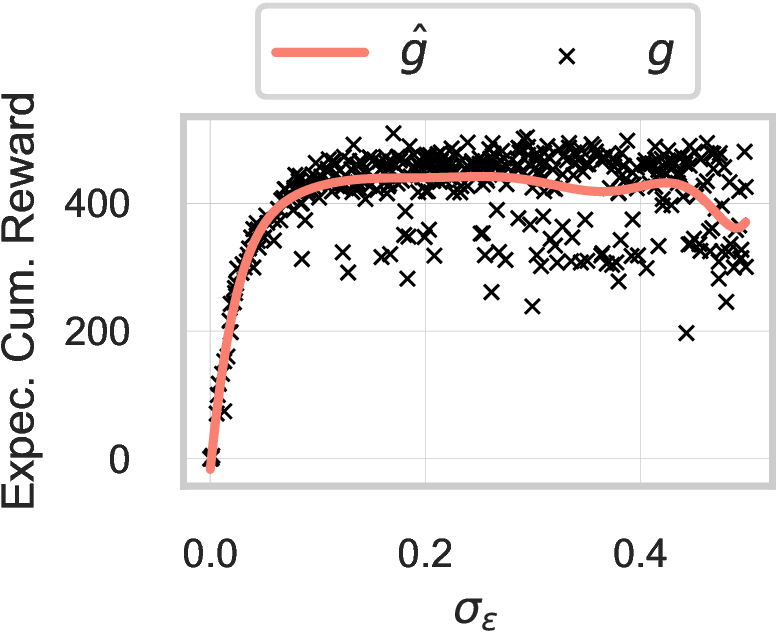}\hspace{0.1mm}
\includegraphics[width=0.420\columnwidth]{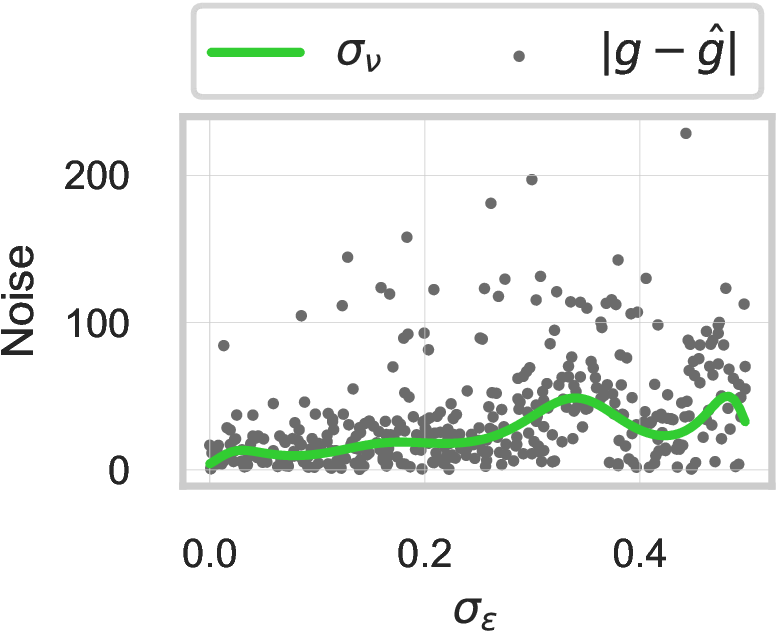}
\caption{Expected reward function for the Half-Cheetah task in (a) using a 10-degree polynomial regression estimate $\hat g$, and the fitted noise $\sigma_\nu$ in (b) seen as the green curve.}
\label{fig:fitted}
\end{figure}

Having a parametric form for $k_\obsNoise$, we learn noise parameters separately in a two-stage regression problem. With a set of sample inputs $\parameters_i \in \Sspace$, we can approximate the expected reward $\hat g$ as in \autoref{fig:fitted} using a flexible generalised linear regression model $\vec{\rho}: \ParamSpace \to \R^\nFeatures$ by: 
\begin{equation}
\begin{split}
\hat{g}(\parameters) \approx \vec{\alpha}^\transpose \vec{\rho}(\parameters)~.
\end{split}
\label{eq:reward-approx}
\end{equation}
With the estimate $\hat g$, we then fit the residuals  $|g(\parameters) - \hat{g}(\parameters)|$ with \eqref{eq:noise-model} as a regression problem that leads to finding $\sigma_\nu$.

\section{Related Work}
\label{sec:related}

The importance of physics and robotic simulation tools is undeniable because they make it possible to generate large amounts of data for learning models \cite{Todorov2012, Cranmer2020}. However, there are two main problems with this approach. Simulators do not fully replicate real hardware behaviour and complex environments, and there is a limit on the amount of data to collect due to resource limitations. These problems are related to the reality gap, or sim-to-real problem \cite{Peng2017, Muratore2021}, and they become noticeable when transferring policies learned in simulators to real systems.

\begin{subsection}{Model-based Methods and Domain Randomisation}

In model-based methods, a dynamics or transition model is fit by maximising the likelihood of a set of collected trajectories. This optimisation does not always correlate with optimising the system reward, as demonstrated in \cite{lambert20a}.

As a way to deal with the reality gap, rather than learning a policy on a single simulated environment, domain randomisation randomises the simulator to expose the policy to a wide range of environments at training time \cite{Tobin2017}. In the literature, dynamics randomisation has been applied to find approximations for robust dynamics in common control and robotics problems. For example, there are generalised transition models that address environment contexts where a robot's dynamics could change due to some robotic part malfunctioning \cite{Lee2020}. Other approaches propose inferring simulation parameters based on data instead of uniform parameter randomisation \cite{Peng2017, Ramos2019}.

Bayesian optimisation has also been applied to adapt domain parameter distributions during learning, improving sim-to-real transfer for classic control problems. \cite{Muratore2021, Oliveira2021}. In particular, the approach in \cite{Muratore2021} also jointly optimises controller parameters. However, they rely on external algorithms \cite{Schulman2017} to optimise control policies and do not account for heteroscedasticity in the reward signal.

\end{subsection}

\begin{subsection}{Stochastic MPC}

Stochastic model predictive control (MPC) has several variants applied from steady-state control to path planning \cite{Williams2018a} to robotics \cite{Carron2019}. This is due to their capacity to deal with highly non-linear dynamics and optimise non-convex reward or cost functions. Moreover, because of the similarities between model predictive control and model-based reinforcement learning \cite{Gorges2017}, approaches exploiting both types of methods have been proposed. In \cite{Carron2019}, for instance, stochastic MPC together with Gaussian processes allows the system to adapt to disturbances using the GP to model uncertainty where little data is available for a robotic arm. 

\end{subsection}

\begin{subsection}{Heteroscedastic Bayesian optimisation}

Bayesian optimisation has been applied to model selection and high-dimensional hyper-parameter tuning, where BO methods form some of the main sample-efficient, derivative-free optimisation methods used in robotics \cite{Oliveira2018}. There have been hyper-parameter tuning approaches for MPC using BO, which optimise back-off terms for thermal energy storage \cite{Lu}, showing considerably optimal terms in a number of simulations. Other approaches propose MPC as a performance-driven controller by optimising design parameters with BO \cite{Piga2019}. Another method uses heteroscedastic BO for MPC \cite{Guzman2020}, which optimises MPC hyper-parameters with BO. The difference with this last one is that we add transition model parameter distributions to the search space.

BO infers samples modelled by a Gaussian process, usually defined with homoscedastic \iid observation noise \cite{Rasmussen2006}. The problem resides when modelling heteroscedastic noise. There are two main ways to do it: interpreting the heteroscedasticity using a separate model or integrating the heteroscedasticity within the GP \cite{Liu2020}. 

One approach for handling heteroscedasticity is to use a separate model for the noise, which could be less expensive for big data because the system would need to fit a noise function beforehand \cite{Pereira2014}. Other approaches are useful for real-time systems by optimising the noise in real-time. The approach in \cite{Liu2020} adds a diagonal term to the GP covariance matrix to model-independent, non-identically distributed noise, which makes inference analytically intractable, requiring approximate solutions. Another approach is to model input-dependent noise variance with a separate GP and then perform variational inference to estimate an approximate posterior GP \cite{Lazaro-Gredilla2011}. That approach has been applied to control learning \cite{Kuindersma2012}. However, in addition to the extra computational cost of variational inference, the method in \cite{Kuindersma2012} does not evaluate uncertainty in the dynamics.

\end{subsection}
\section{Experiments}
\label{sec:experiments}

We evaluate the performance of the proposed adaptive model predictive control by solving simulated control and robotic tasks. In the first subsection, we specify the tasks and then the dynamics model and controller settings used in the experiments. Next, we evaluate the approach using optimisation methods to support the use of adaptive MPC. Finally, we present results with robotic tasks. More details can be found in the video submitted with the paper.

\subsection{Optimisation Variables}

We evaluated adaptive MPC in OpenAI\footnote{OpenAI Gym: \url{https://gym.openai.com}} benchmark tasks: Cartpole, Pendulum, Half-Cheetah, Reacher, and Fetchreach with dense reward functions. The same functions taken from \cite{Guzman2020} are applied to compute instant costs for stochastic MPC. 

First of all, different tasks have different constraints for model parameters. Some parameters have to be positive, like the mass and length, and others can be defined only within a specific range. We considered using a probability distribution with positive support for $\boldsymbol{\theta}$. A beta distribution has support $[0, 1]$, which is inconvenient since one has to scale sampled model parameters if values outside of this range are needed. We also need to specify ranges for distribution parameters $\boldsymbol{\psi}$, e.g. if we define model parameters as gamma distributed, we need to specify ranges for $\alpha$ and $\beta$. We optimise the distribution mean $\mu$ and standard deviation $\sigma$ and then derive $\alpha$ and $\beta$. We use gamma distributions for the simplicity of such conversion. We define each model parameter as gamma-distributed: $\theta \sim \operatorname{Gamma}(\alpha, \beta)$, where $\alpha=\frac{\mu^2}{\sigma^2}$ and $\beta=\frac{\mu}{\sigma^2}$. 

Next, we define ranges for the parameterisation $\boldsymbol{\psi}$ and the controller hyper-parameters $\boldsymbol{\phi} = \{\lambda, \sigma_\epsilon\}$. These ranges were found by narrowing down large enough intervals until noticeable optimal regions are found. Then, we use heteroscedastic BO, to perform episodic cumulative reward optimisation from \autoref{eq:optproblem} where we jointly optimise $\boldsymbol{\psi}$ and $\boldsymbol{\phi}$.


\subsection{Experiments on Classic Control Tasks}

We aim to find optimal hyper-parameter settings $\vec\phi$ together with model distribution parameters $\vec\psi$. For Cartpole and Pendulum, $\mu_l$ and $\mu_m$ are the mean pole length and mass. For Half-Cheetah and Reacher, we consider the scaling factor $\kappa$ as a random variable with mean and standard deviation denoted by subscripts. $\kappa$ multiplies the model parameters, e.g. $\kappa$ with mean $\mu=1.0$ and $\sigma \to 0$ means that the model parameter corresponds exactly to its true value. $\kappa_m$ and $\kappa_d$ are the scaling factors of the masses and damping ratios for Half-Cheetah and Fetchreach, and $\kappa_{d1}$ and $\kappa_{d2}$ are scaling factors for two Reacher damping ratios.

\begin{center}
\begin{table}[h]
    \centering
    \setlength{\tabcolsep}{0.34em} 
    \renewcommand{\arraystretch}{1.1} 
    \begin{tabular}{|c|c|c|c|c|c|}
        \hline
        \rowcolor{gainsboro}
        Problem     &  $n_e$ & $T$ & $M$ & \multicolumn{2}{c|}{Distribution parameter range} \\
        \hline                      
        Cartpole     & 40 & 10  & 250 & \shortstack{$\mu_l \in {[}0.2, 1.5{]}$\\ $\mu_{m} \in {[}0.1, 1.5{]}$} & \shortstack{$\sigma_l \in  {[}1e-5, 0.1{]}$  \\ $\sigma_m \in  {[}1e-5, 0.1{]}$}      \\
        \hline
        Pendulum    & 15 & 20 &  400  & \shortstack{$\mu_l \in {[}0.2, 2.0{]}$\\ $\mu_{m} \in {[}0.2, 2.0{]}$} & \shortstack{$\sigma_l \in  {[}1e-5, 0.1{]}$  \\ $\sigma_m \in  {[}1e-5, 0.1{]}$}      \\
        \hline
        Half-Cheetah  & 25 & 14 & 10 & \shortstack{$\kappa_{m,\mu} \in {[}0.6, 1.2{]}$ \\ $\kappa_{d,\mu} \in {[}0.6, 1.4{]}$} & \shortstack{$\kappa_{m,\sigma} \in  {[}1e-5, 0.1{]}$ \\ $\kappa_{m,\sigma} \in  {[}1e-5, 0.1{]}$}     \\
        \hline
        Reacher  &  20   &  12 & 18 & \shortstack{$\kappa_{d1,\mu} \in {[}0.1, 8.0{]}$ \\ $\kappa_{d2,\mu} \in {[}0.1, 8.0{]}$} & \shortstack{$\kappa_{d1,\sigma} \in  {[}0.001, 0.6{]}$ \\ $\kappa_{d2,\sigma} \in  {[}0.001, 0.6{]}$}     \\
        \hline
        Fetchreach  & 120 & 3 & 12 & $\kappa_{d,\mu} \in {[}1.0, 50.0{]}$ & $\kappa_{d,\sigma} \in  {[}0.001, 0.6{]}$     \\
        \hline
    \end{tabular}
    \caption{Ranges of parameters and scaling factors.}
    \label{tab:pranges}
\end{table}
\end{center}

\begin{figure}[ht]
\centering
\includegraphics[width=0.165\textwidth]{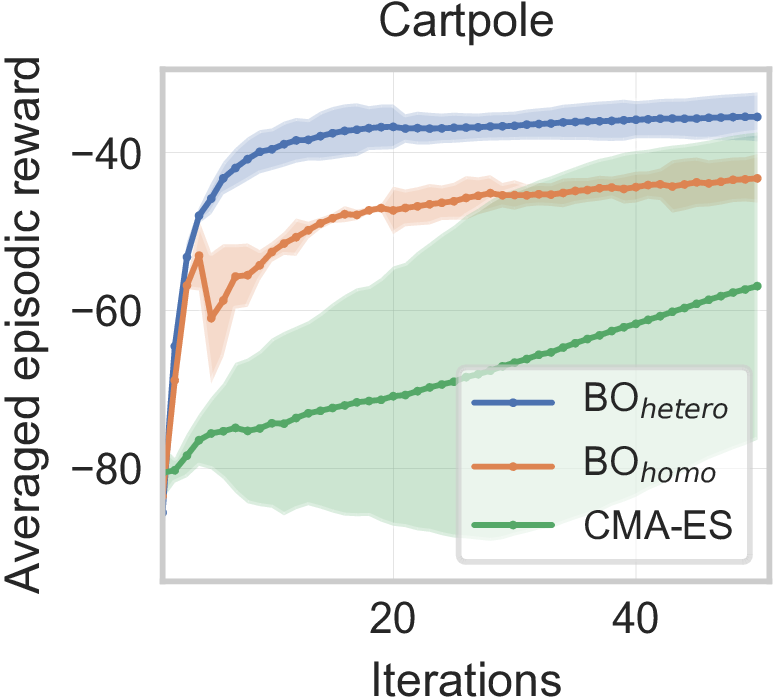}
\includegraphics[width=0.15\textwidth]{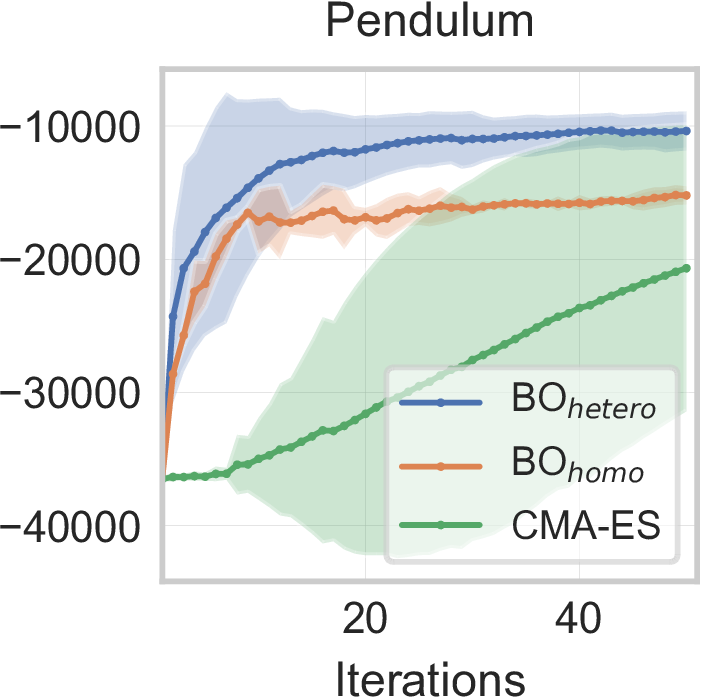}
\includegraphics[width=0.15\textwidth]{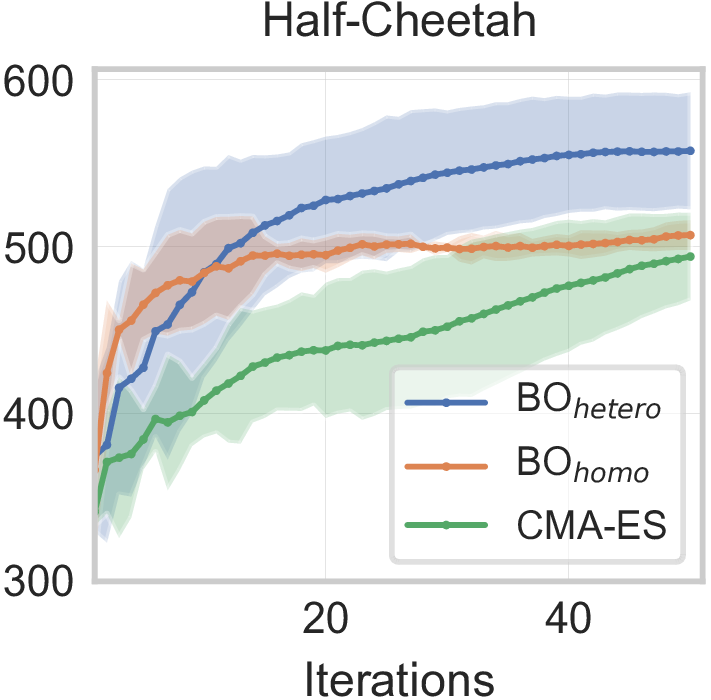}\\
\vspace{1mm}
\includegraphics[width=0.16\textwidth]{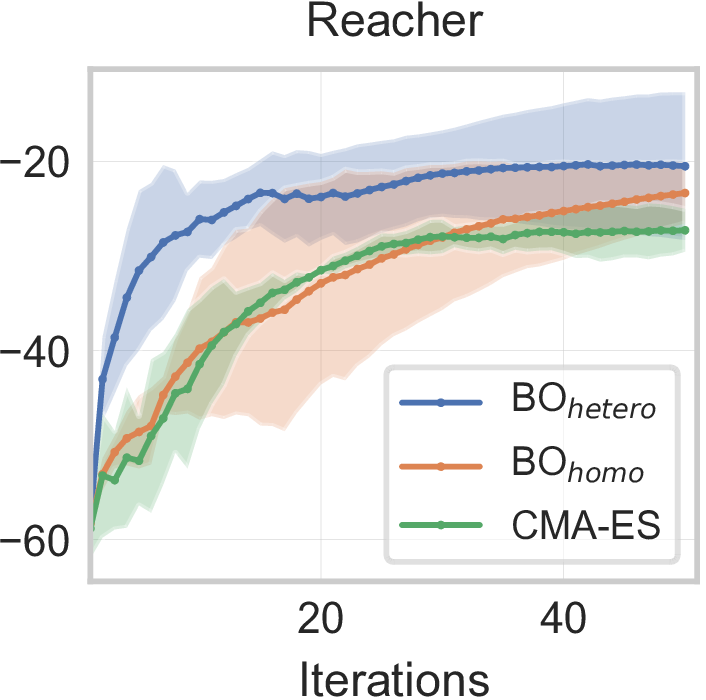}
\includegraphics[width=0.15\textwidth]{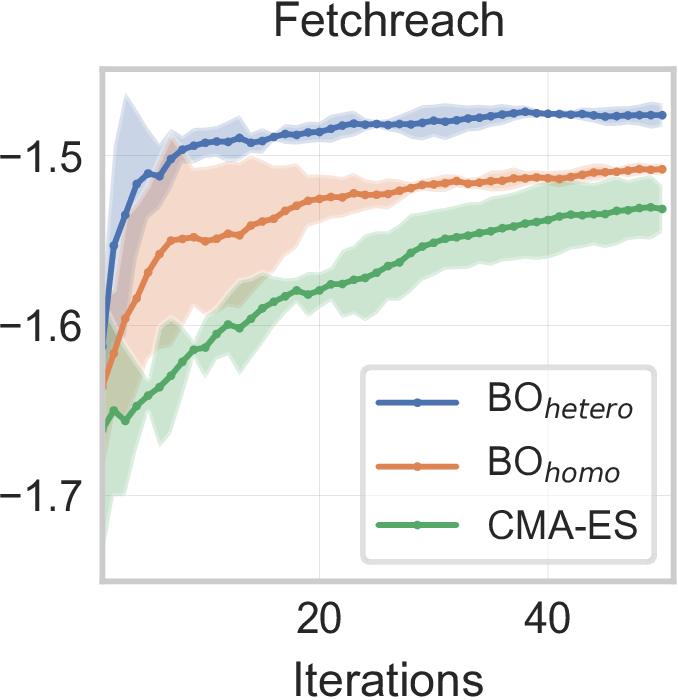}
\caption{Averaged cumulative rewards per iteration. The shaded areas denote two standard deviations. Each method started at a point with minimum expected cumulative reward.}
\label{fig:rewardcurves}
\end{figure}

We use true parameter values of 1.0 and search spaces within $\lambda \in {[}1e-5, 2.5{]}$, $\sigma_\epsilon \in {[}1e-5, 4.0{]}$ for the controller hyper-parameters and distribution parameters with ranges shown in \autoref{tab:pranges} with $n_s=200$ for every task. Results in \autoref{fig:rewardcurves} show averaged cumulative rewards for 50 iterations by running both BO versions, and covariance matrix adaptation evolution strategy (CMA-ES) \cite{Arnold2010}, which was configured according to \cite{Guzman2020}. Each BO method had GP kernel hyper-parameters and noise model globally optimised via maximum likelihood estimation from a batch of 150 observations. We used an exponential kernel for both. The maximum likelihood optimisation for finding optimal GP kernel settings is done using L-BFGS-B with random initialisations. For both BO versions, we used the exponential kernel $k(x, x') =  \sigma_n^2\exp\left(-\frac{(x - x')^2}{2\ell^2}\right)$. We set $\Omega := \{\sigma_\nu,\sigma_n,\ell\}$ for homoscedastic BO and $\Omega := \{z, \sigma_n,\ell\}$ for heteroscedastic BO. For both BO versions, we used a UCB acquisition function \eqref{eq:ucb_problem} with $\delta=2.0$ optimised with the same global L-BFGS-B procedure.

We can see that heteroscedastic BO can outperform the others due to the noise nature. Each optimisation starts at points that give a minimum reward. On average, adaptive MPC with heteroscedastic BO can find optimal $\vec{\psi}$ and $\vec{\phi}$ in fewer iterations and, in most cases, with less variance.

\subsection{Adapted Parameter Distributions}

\begin{figure}[ht]
    \centering
    \includegraphics[width=0.45\textwidth]{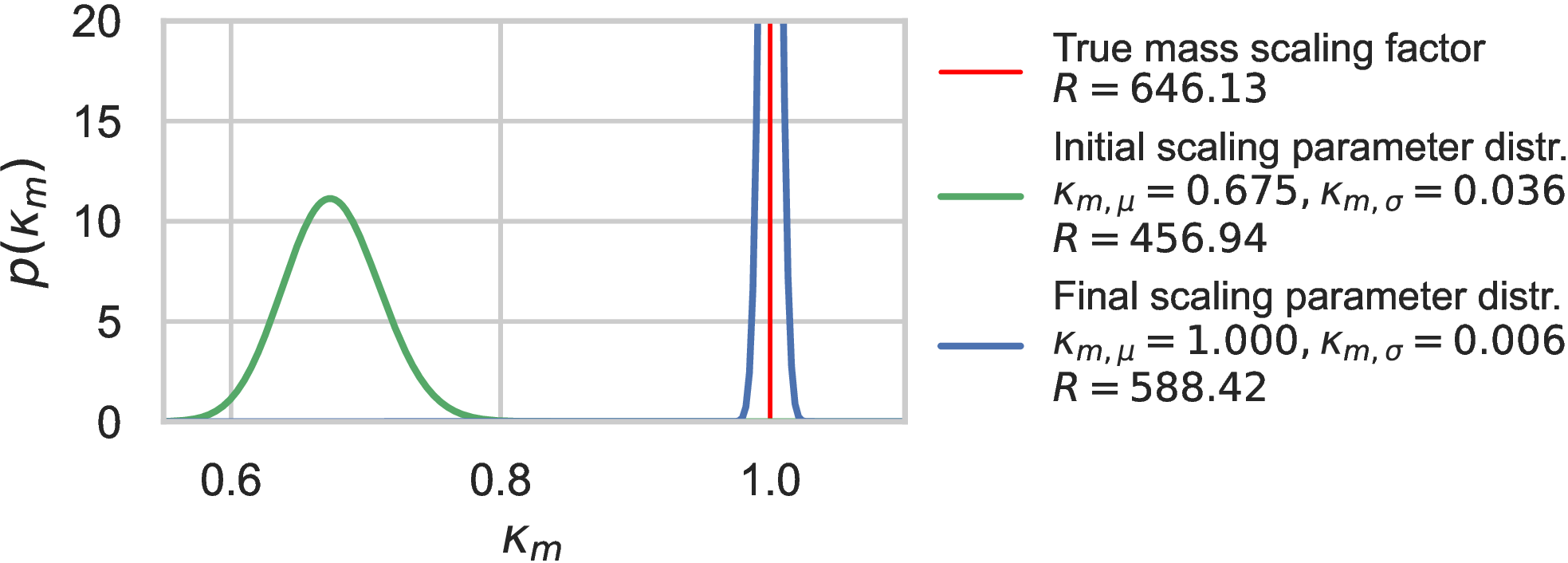}
    \caption{Mass parameter scaling factor $\kappa_{m}$ for Half-Cheetah.} \label{fig:distright}
\end{figure}

For Half-Cheetah, the optimised distribution of the scaling factor of the mass $\kappa_{m}$ yields a reward of $701.29$, which is higher than the reward obtained with true value. This can be explained because we optimise the reward for the controller and not for system identification. In \autoref{fig:distright}, we show the cumulative reward of using the true parameters and the optimised distribution. The optimal parameter distribution estimated shows the approximation to the true value.

\begin{figure}[ht]
    \centering
    \includegraphics[width=0.45\textwidth]{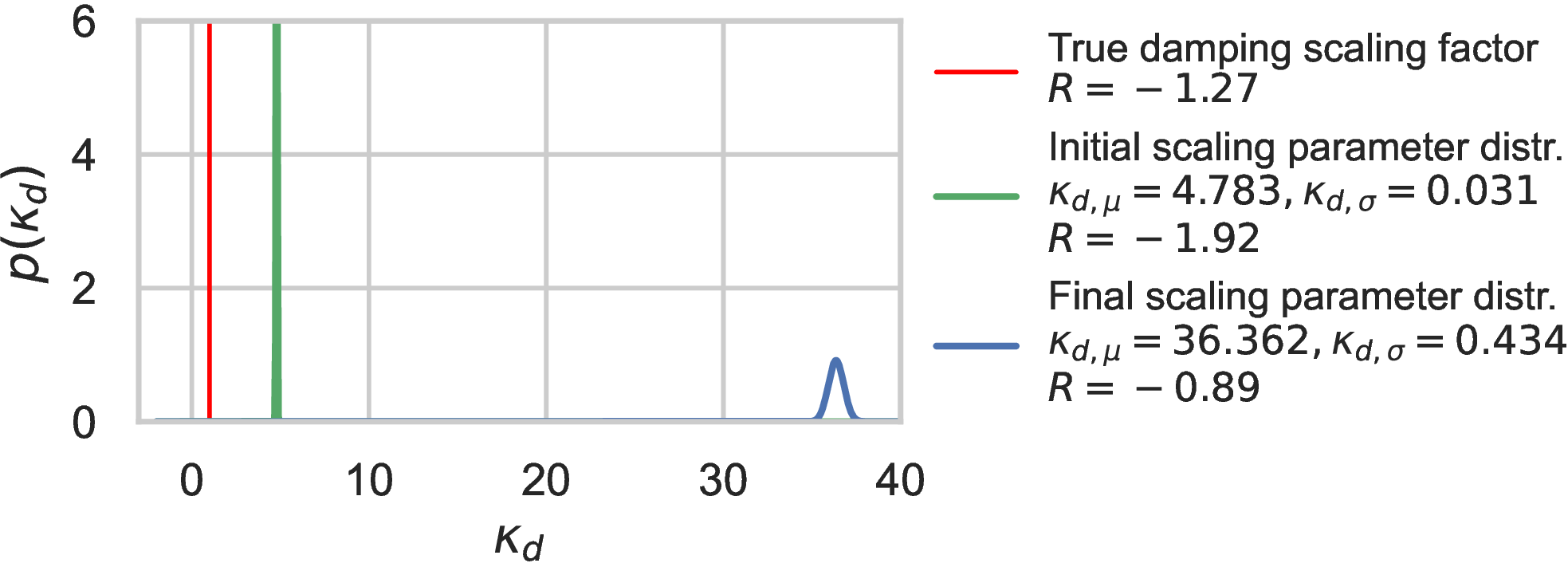}
    \caption{Optimised scaling factor $\kappa_d$ for Fetchreach.} \label{fig:distrfetchreach}
\end{figure}

In the case of Fetchreach, the damping ratio plays a major role. We found out that even wrong dynamics model parameters can lead to better performance. In \autoref{fig:distrfetchreach}, we start at an initial parameter distribution denoted in blue for the damping ratio $\kappa_d$ that gives a maximum reward found via random search. Then, we use adaptive MPC with heteroscedastic BO to optimise the parameter distribution. We found that a high damping ratio allows faster movements, improving reward performance with the true model. That is because, for an overdamped model, the controller would be prone to apply higher torques by choosing a larger control variance $\sigma_\epsilon$. However, that leads to uncertain arm movements when the gripper is close to the target. 

\subsection{Experiments on a Robotic Simulator}

We used a simulated PANDA robotic arm\footnote{IssacGym: \url{https://developer.nvidia.com/isaac-gym}} with the task of reaching a yellow target with a gripper in a single-obstacle environment using the MPPI-based motion planning framework from \cite{Bhardwaj-CoRL-21}. We define $n_e=10$, $n_s=480$, and the fixed hyper-parameters $T=20$, $M=150$, and $\sigma_\epsilon=0.5$.

\begin{figure}[h]
    \centering
    \subfloat[Robot task]{\includegraphics[width=0.45\columnwidth]{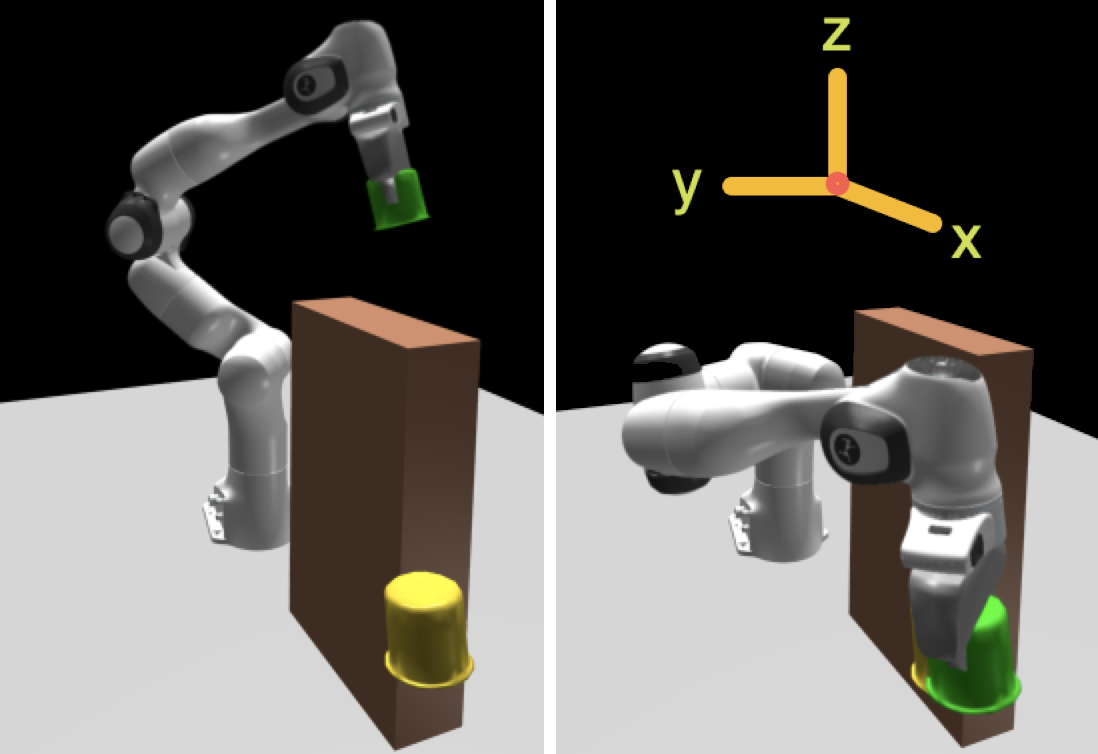}\label{fig:franka}}
    \hspace{3mm}
    \subfloat[Reward optimisation]{\includegraphics[width=0.4\columnwidth]{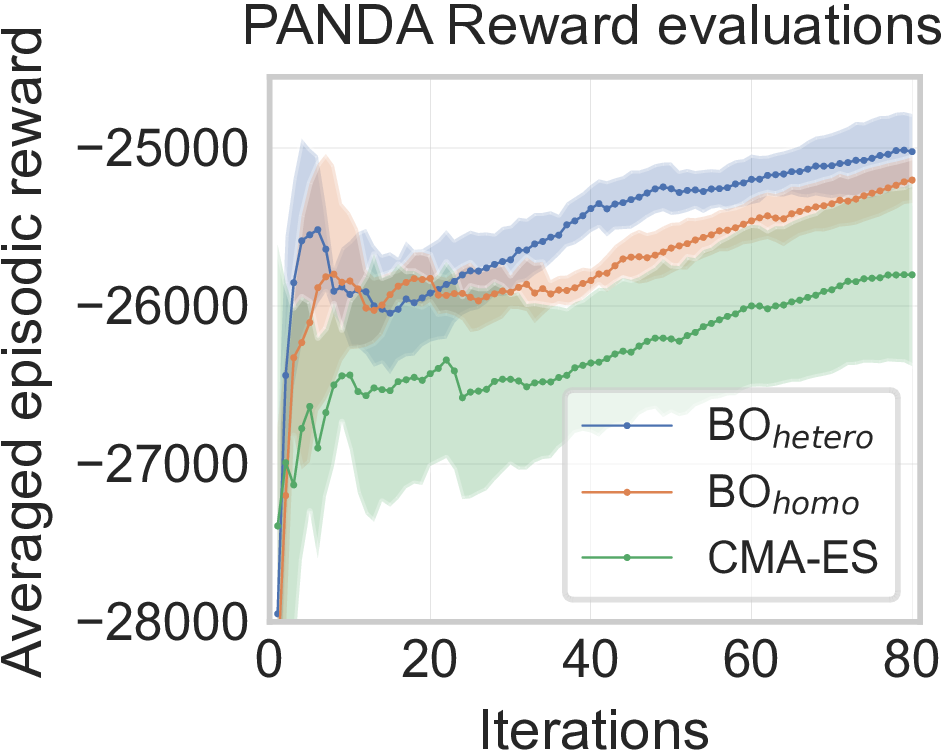}\label{fig:franka_rewards}}\\
    \subfloat[Heteroscedastic behavior]{\includegraphics[width=0.5\columnwidth]{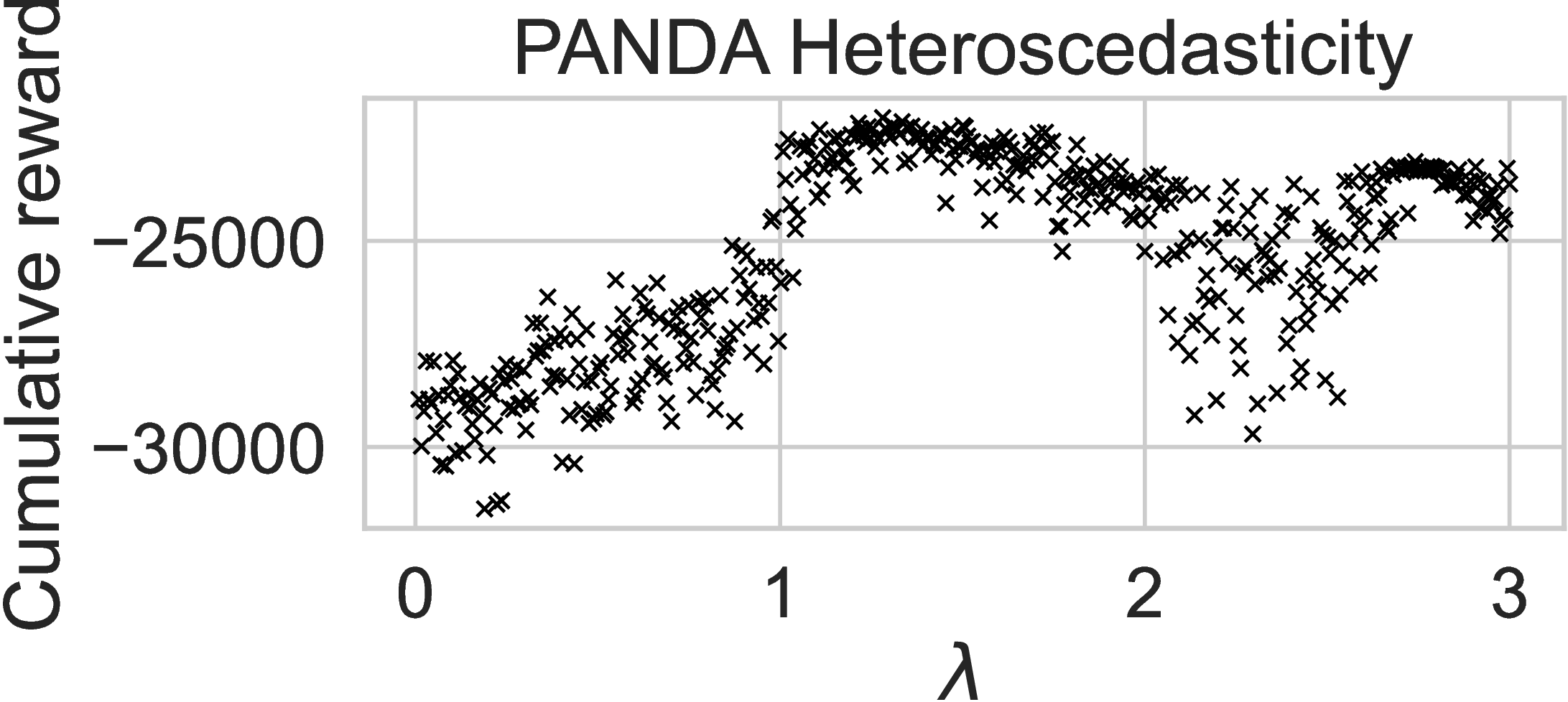}\label{fig:franka_heteroscedastic}}
    \caption{Experiments with the PANDA simulator.}
    \label{fig:robot}
\end{figure}

Trajectory evaluations are done on the GPU, which helped overcome efficiency issues of MPC. Then, for simplicity, the task only has one obstacle as in \autoref{fig:franka}. In this case, we adapt the controller hyper-parameter $\phi=\lambda \in {[}0.01, 2.0{]}$ to the environment by defining distribution-based length $x$, width $y$ and height $z$ defined by $\psi = \{x_{\mu}, x_{\sigma}, y_{\mu}, y_{\sigma}, z_{\mu}, z_{\sigma}\}$, and with true values (0.3, 0.1, 0.6) respectively. Their search space is limited to 0.02 higher than each true size.

In \autoref{fig:franka_rewards}, we compare the averaged cumulative reward against the number of iterations with each method after running 80 iterations. The heteroscedastic behaviour of $\lambda$ is shown in \autoref{fig:franka_heteroscedastic}, and that is exploited by BO$_{hetero}$.

\subsection{Experiments on a Real Robot}

We configured a reaching task using the Jaco 2 robot shown in \autoref{fig:jaco} (left). We define $n_e=8$, $n_s=1$, and the fixed hyper-parameters $T=1$, $M=15000$, and $\sigma_\epsilon=1$. The action space is composed of the 6 joint angles. The objective is to reach a fixed red target from Jaco's predefined base position. The arm length $d_1$ and the front arm length $d_2$ with true values (0.4100, 0.2073) are defined as randomised variables with $\psi = \{d_{1,\mu}, d_{1,\sigma}, d_{2,\mu}, d_{2,\sigma}\}$, and we adapt them to $\phi=\{\lambda\}$ as with the PANDA robot. We define intervals of $\pm 0.10$ the true lengths, and standard deviations $d_{1,\sigma}, d_{2,\sigma} \in (0.001, 0.03)$. The reward is computed as the negative distance from the gripper to the target.

\begin{figure}[h]
    \centering
    \includegraphics[width=0.34\columnwidth]{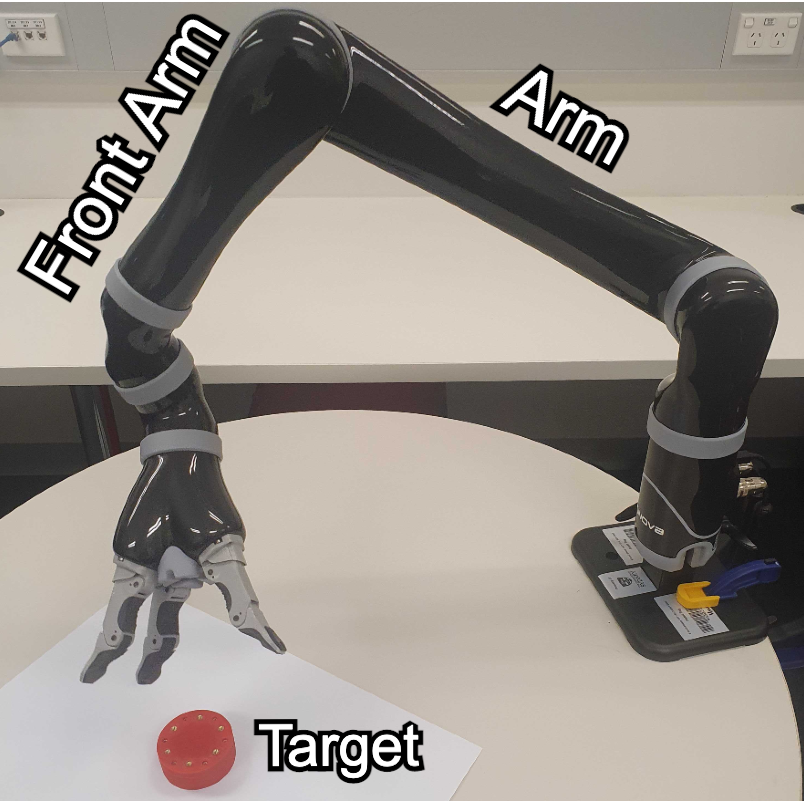}
    \hspace{3mm}
    \includegraphics[width=0.20\textwidth]{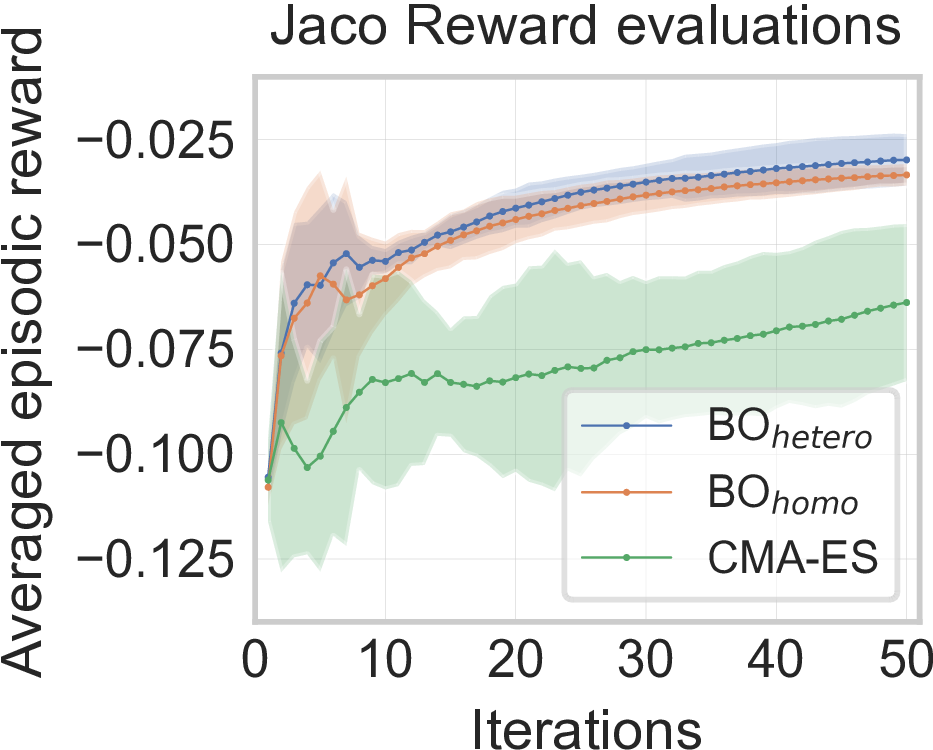}
    \caption{Jaco robot (left) and reward optimisation (right).} \label{fig:jaco}
\end{figure}

\begin{figure}[h]
    \centering
    \includegraphics[width=0.48\columnwidth]{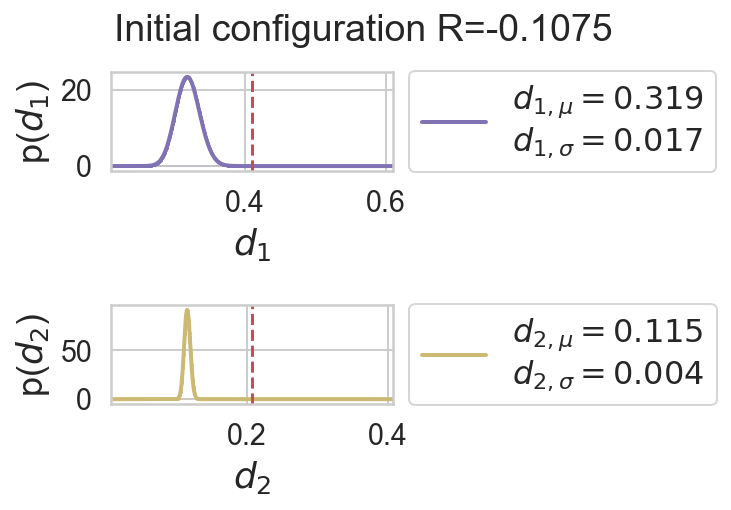}    \includegraphics[width=0.48\columnwidth]{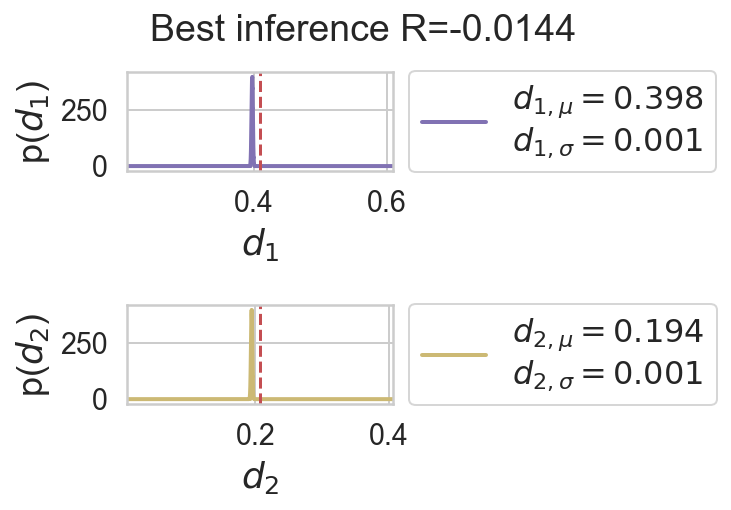}
    \caption{Initial distribution (left) and best inference after 50 iterations (right).} \label{fig:jaco}
\end{figure}

\autoref{fig:jaco} shows the best approximation by BO$_{hetero}$ of the length variables to their true values after 50 iterations as shown in \autoref{fig:jaco} (right). Optimal values of those variables should reach point estimates, but optimising point estimates is out of the scope of this work. Similar optimal distributions are inferred by both BO versions in this case.

\section{Conclusions}

This work addresses the problem of optimising the performance of stochastic MPC in uncertain environments. We presented a BO framework that simultaneously optimises the internal dynamics model and the controller itself. In this context, we verified the heteroscedasticity in controller hyper-parameters and dynamics model parameters for control tasks and robotic tasks since heteroscedastic BO was able to do as well or better than the traditional BO. Experiments were run with two versions of BO and a non-BO method for reward optimisation, and results showed that having parameter and object dimension distributions can lead to improved performance in a few iterations. For future work,  an approach could be to use a  non-GP surrogate model for  BO,  such as a Bayesian deep neural network.  Another point would be to obtain theoretical results on the effects of heteroscedasticity and on the performance of the framework. 


\bibliographystyle{IEEEtran}
\bibliography{rl_bo_references}
\end{document}